\documentclass[conference]{IEEEtran}
\IEEEoverridecommandlockouts
\usepackage{cite}
\usepackage{amsmath,amssymb,amsfonts}
\usepackage{algorithmic}
\usepackage{graphicx}
\usepackage{textcomp}
\usepackage{url}
\def\BibTeX{{\rm B\kern-.05em{\sc i\kern-.025em b}\kern-.08em
    T\kern-.1667em\lower.7ex\hbox{E}\kern-.125emX}}
\begin{document}

\title{Classification of fNIRS Data Under Uncertainty: A Bayesian Neural Network Approach\\}

\author{

\IEEEauthorblockN{Talha Siddique}
\IEEEauthorblockA{\textit{Electrical and Computer Engineering} \\
\textit{University of New Hampshire}\\
Durham NH, USA \\
ts1121@wildcats.unh.edu}
\and
\IEEEauthorblockN{Md Shaad Mahmud}
\IEEEauthorblockA{\textit{Electrical and Computer Engineering} \\
\textit{University of New Hampshire}\\
Durham NH, USA \\
mdshaad.mahmud@unh.edu}
}

\maketitle

\begin{abstract}
Functional Near-Infrared Spectroscopy (fNIRS) is a non-invasive form of Brain-Computer Interface (BCI). It is used for the imaging of brain hemodynamics and has gained popularity due to the certain pros it poses over other similar technologies. The overall functionalities encompass the capture, processing and classification of brain signals. Since hemodynamic responses are contaminated by physiological noises, several methods have been implemented in the past literature to classify the responses in focus from the unwanted ones. However, the methods, thus far does not take into consideration the uncertainty in the data or model parameters. In this paper, we use a Bayesian Neural Network (BNN) to carry out a binary classification on an open-access dataset, consisting of unilateral finger tapping (left- and right-hand finger tapping). A BNN uses Bayesian statistics to assign a probability distribution to the network weights instead of a point estimate. In this way, it takes data and model uncertainty into consideration while carrying out the classification. We used Variational Inference (VI) to train our model. Our model produced an overall classification accuracy of 86.44\% over 30 volunteers. We illustrated how the evidence lower bound (ELBO) function of the model converges over iterations. We further illustrated the uncertainty that is inherent during the sampling of the posterior distribution of the weights. We also generated a ROC curve for our BNN classifier using test data from a single volunteer and our model has an AUC score of 0.855.
\end{abstract}


\section{Introduction}
A Brain-Computer Interface (BCI) is a communication system that allows brain activities to control external devices or even computers\cite{10.3389/fnhum.2015.00003}. The overall functionalities of the BCI encompass the capturing of brain signals, processing the received signals, classifying them, and utilizing these signals for control. Functional Near-Infrared Spectroscopy (fNIRS) is a non-invasive form of BCI which is used for functional monitoring and imaging of brain hemodynamics \cite{10.3389/fnhum.2015.00003}. The configuration of the system consists of light sources and detectors that are placed on the scalp and two wavelengths of near-infrared light is transmitted through the top layer of the cerebral cortex \cite{10.3389/fnhum.2015.00003}. Near-infrared light between 650 and 950 nm can pass through skin, bone and water, but absorbed by both oxygenated-hemoglobin (HbO) and deoxygenated-hemoglobin (HbR) \cite{10.3389/fnhum.2015.00003}. Due to this, the relative changes in the concentration of HbO and HbR can be calculated from changes in the reflected dual-wavelength light using the modified Beer-Lambert Law. Compared to the other BCI counterparts, fNIRS is safer, less costly, portable, and provides a higher temporal resolution \cite{10.3389/fnhum.2015.00003}. Because of this fNIRS has gained popularity as a research and clinical tool.

Hemodynamic responses are contaminated by physiological noises and therefore studies have been conducted to classify the task related responses from unwanted responses \cite{bak2019open}. The literature consists of studies where the BCI data set was band-pass filtered and methods such as Linear Discriminant Analysis was used to extract the feature set for the classification prediction \cite{10.3389/fnhum.2015.00003}. Other types of machine learning algorithms have been used, at times in conjunction with filtering methods. For example, the fNIRS data set was band-pass filtered and a ternary classification was carried out using Support Vector Machine (SVM) \cite{10.3389/fnhum.2015.00003}; Maximum-likelihood classifiers specific to the task and subject were developed using hidden Markov models on fNIRS dataset \cite{10.3389/fnhum.2015.00003}.

However, the literature lacks models that take into account the uncertainty in the BCI/fNIRS data or the model and analyze the effects it might have on the prediction. Generally, there are two types of major uncertainty that can be modelled- epistemic uncertainty and aleatoric uncertainty \cite{kwon2020uncertainty}. Epistemic uncertainty accounts for the uncertainty in the model parameter. This kind of uncertainty can be reduced with more data. Therefore, epistemic uncertainty is higher in the regions where there is little or no training data and lower in the regions of more training data. Aleatoric uncertainty is related to the noise inherent in the training dataset itself. Such uncertainty cannot be reduced if we get more data.

To address the aforementioned gap in BCI/fNIRS literature, we used a Bayesian Neural Network (BNN) to carry out a binary classification using an open-access fNIRS dataset. The BNN allows us to carry out the classification process while capturing both the epistemic and aleatoric uncertainty.

\begin{figure}[htp]
    \centering
    \includegraphics[width=8.5cm,height=6cm]{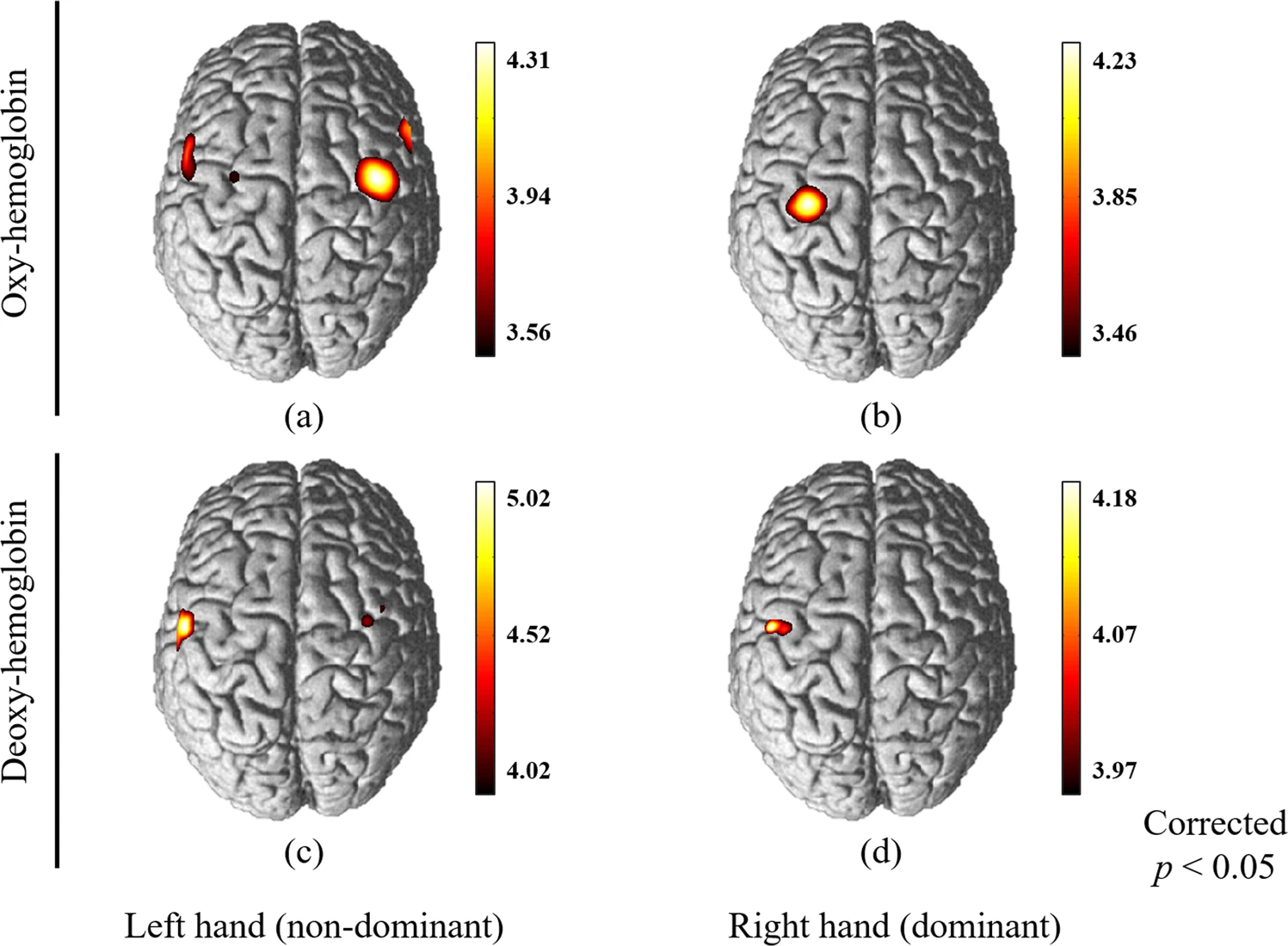}
    \caption{Cortical activation patterns of chopstick tasks. (a) cortical mapping based on changes in oxyhemoglobin
using left hand (non-dominant), (b) cortical mapping based on changes in oxy-hemoglobin using
right hand (dominant) movement. (c) cortical mapping based on changes in oxy-hemoglobin using left hand
(non-dominant) and (d) cortical mapping based on changes in oxy-hemoglobin using right hand (dominant)
movement) \cite{lee2019difference}}
    \label{fig_1}
\end{figure}

\section{Methodology}

\subsection{Data Acquisition and Processing}

\begin{figure}[htp]
    \centering
    \includegraphics[width=4.5cm,height=5cm]{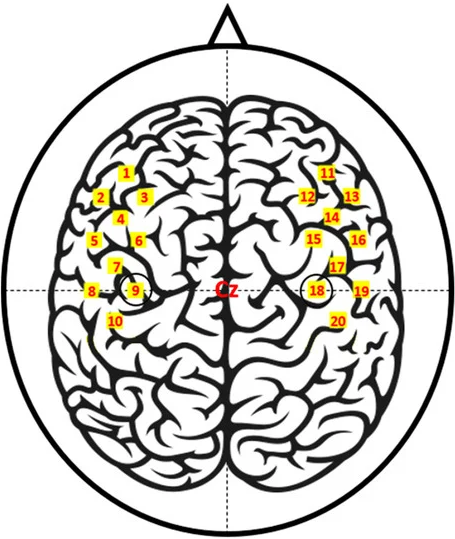}
    \caption{fNIRS channel locations \cite{boas2014twenty}}
    \label{fig_2}
\end{figure}

In our paper, an open-access fNIRS dataset consisting of unilateral finger and foot tapping was used \cite{bak2019open}. In their study \cite{bak2019open}, a total of 30 volunteers participated (29 right-handed; 17 males; 23.4-2.5 years old). The volunteers did not have any pre-existing condition which could influence their study. A three-wavelength continuous-time multi-channel fNIRS system consisting of eight light sources (Tx) and eight detectors (Rx) were used to record the fNIRS dataset. Fig-\ref{fig_2} shows an example illustration of the placement of the Tx and Rx. On the left hemisphere around the C3 region, four of each Tx and Rx were placed and the remaining sources and detectors were placed on the right hemisphere around the C4 region.  

In their experiment, the volunteers sat in front of a 27 inch LED monitor which displayed the triggers for three distinct tasks- Right Finger Tapping (RFT), Left Finger Tapping (LFT) and Foot Tapping (FT). Each task consisted of 25 trials. A single trial consisted of an introductory period of 2 s and a task period of 10 s, followed by an inter-trial break of 17–19 s. For the finger tapping, the volunteers tapped their thumbs with the other fingers, one by one, starting from the index finger and then repeated the movement in reverse order. The volunteers performed the tapping at a constant rate of 2 Hz. As for the foot tapping, the volunteers tapped the foot on their dominant side at a constant rate of 1 Hz. 

In the proposed work, the aforementioned data of the 30 volunteers were available to us in a Matlab structure array format. A volunteer’s data consisted of concentration changes of oxygenated/reduced hemoglobin $\Delta$HbO/HbR (cntHb), trigger (mrk), and fNIRS channel information (mnt). We used the BBCI toolbox, which was recommended by Bak et al. to process the dataset in accordance with the description provided by them. In order to remove the physiological noises and DC offset, a signal processing step was carried out, where the concentration changes of oxygenated/reduced hemoglobin were band-pass filtered through a zero-order. The band pass filter was implemented by the third-order Butterworth filter with a passband of 0.01–0.1 Hz. Relative to the task onset (i.e., 0 s), the $\Delta$HbO/R values were further segmented into epochs ranging from -2 to 28 s. These epochs were then subjected to a baseline correction to subtract the average value within the reference interval of -1-0s.

The feature extraction process was carried out in accordance with their description, where three time windows were employed in the ranges of 0–5 s, 5–10 s, and 10–15 s within epochs to compute the average $\Delta$HbO/R for each of the 20 channels. The resulting dataset consisted of the task labels along with their corresponding 120 epoch segments. Since in our study, we carried a binary classification, we only considered the data for RFT and LFT.

\subsection{Model Overview}

In our study, we used a Bayesian Neural Network (BNN) to carry out a binary classification between the two tasks- RFT and LFT, while taking data uncertainty into consideration. In the case of a classical Neural Network (NN), the weights are a point estimate. However, in BNN, using Bayesian statistics, a probability distribution is assigned as the weights. The probability distributions capture the uncertainty in the data. We train our BNN using Variational Inference (VI) which is a method to approximate probability densities through optimization. VI differs from other classical methods like Markov Chain Monte Carlo (MCMC) in that it learns the parameters of these distributions instead of the weights directly.

Our BNN can be viewed as probabilistic model $p(y|x,w)$. Here, $y$ is a set of our classes, $y = \{RFT, LFT\}$; $x$ is the set of features (120 epoch segments); $w$ is the weight parameter and $p(y|x,w)$ is a categorical distribution. Given our training dataset $D = \{x, y\}$ we can construct the likelihood function which is a function of the parameter $w$. The likelihood function is as follows:

\begin{equation}\label{eq1}
p(D|w)= \prod p(y |x, w)
\end{equation}

By maximizing the likelihood function, we can get the maximum likelihood estimate (MLE) of $w$, where the objective function is the negative log likelihood. According to Bayes theorem the posterior distribution is proportional to the product of the prior distribution, $p(w)$ and the likelihood $p(D|w)$. However, MLE gives point estimates for parameters due to which the uncertainty over the weights cannot be captured. Therefore, BNN averages predictions over a combination of NN which are weighted by the posterior distribution of the parameter $w$. The mathematical formulation of the posterior predictive distribution is as follows:

\begin{equation}\label{eq2}
p(y|x,D)=\int p(y|x,w)p(w|D) dw
\end{equation}
  
Since the determination of the posterior distribution, $p(w|D)$ is intractable, BNNs can use a variational distribution $q(w|\theta)$ of known functional form to approximate the true posterior distribution. In order to achieve this the Kullback-Leibler (KL) divergence between $q(w|\theta)$ and the true posterior $p(w|D)$ w.r.t. to $\theta$ is minimized \cite{blei2017variational}. The corresponding objective function is as follows:

\begin{multline}\label{eq3}
\text{KL} (q(w|\theta) || p(w|D)) =  \mathbb{E}[\log q(w|\theta)] -  \mathbb{E}[\log p(w)]\\ - \mathbb{E}[\log p(D|w)] + \log p(D)
\end{multline}

Since the KL cannot be calculated, we use the negative of the KL divergence function called the evidence lower bound (ELBO) which does not contain the term $\log p(D)$. Since log $p(D)$ is a constant term, it can be discarded and therefore maximizing the ELBO function is equivalent to minimizing the KL divergence. The functional form of the ELBO function is stated below.

\begin{equation}\label{eq4}
\text{ELBO} (q) =   \mathbb{E}[\log p(w)] + \mathbb{E}[\log p(D|w)] - \mathbb{E}[\log q(w|\theta)]
\end{equation}

Our BNN consisted of 2 hidden layers with 5 neurons and as for the perceptron, we used a sigmoid classifier. We took 70\% of our dataset as training set and the remaining 30\% as test set. The VI method was implemented in our model using Python’s PyMC3 module. We used a Gaussian distribution to represent the prior distribution and the variational posterior distribution. The distribution has the parameter $\theta =(\mu,\sigma)$ where $\mu$ and $\sigma$ is the mean and standard deviation vector respectively of the distribution. Since The $\sigma$ consists of elements of a diagonal covariance matrix, the weights of our BNN can be assumed to be uncorrelated. Since we are parameterizing the model with $\mu$ and $\sigma$, a BNN ends up with a greater number of parameters compared to a classical NN. A training iteration of the model consists of a forward-pass and backward-pass. A single sample is drawn from the variational posterior distribution during a forward pass, which is then used to approximate the ELBO function. During a backward-pass, back propagation is used to calculate gradients of $\mu$ and $\sigma$.

As mentioned before, there are two types of uncertainty- epistemic uncertainty and aleatoric uncertainty. The uncertainty in weights causes epistemic uncertainty. The variational posterior distribution in our model covers the epistemic uncertainty whereas the likelihood function covers the aleatoric uncertainty.

\section{Results and Discussion}

\begin{figure}[htp]
    \centering
    \includegraphics[width=8.5cm,height=6cm]{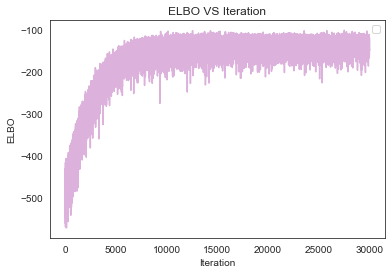}
    \caption{ELBO over iterations}
    \label{fig_3}
\end{figure}

\begin{figure}[htp]
    \centering
    \includegraphics[width=8.5cm,height=6cm]{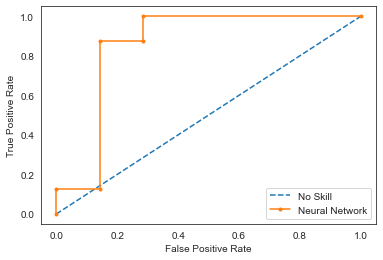}
    \caption{Comparison of Receiver Operator Characteristics between the BNN classifier and a No-Skill classifier, using test data from Volunteer-1}
    \label{fig_4}
\end{figure}

\begin{figure}[htp]
    \centering
    \includegraphics[width=6cm,height=9.5cm]{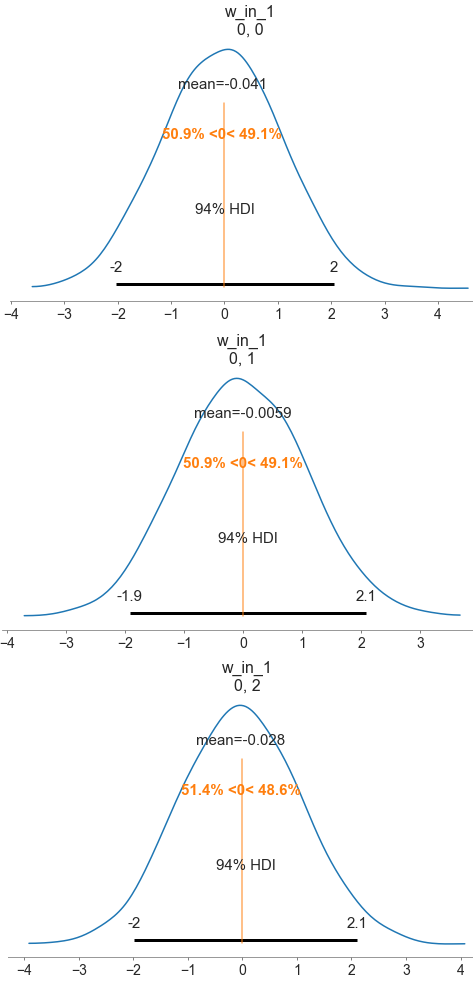}
    \caption{Sampling of a posterior weight distribution over 3 independent iterations}
    \label{fig_5}
\end{figure}

\begin{figure}[htp]
    \centering
    \includegraphics[width=8cm,height=8.5cm]{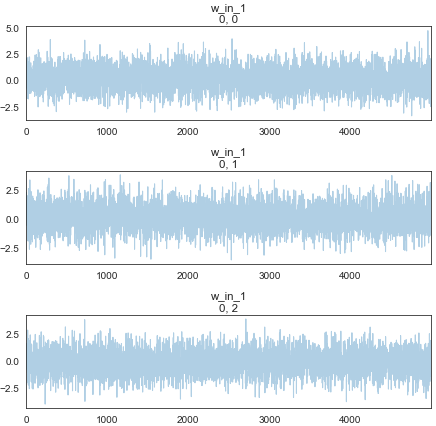}
    \caption{Traceplot showing the sample value of the weight for each iteration. The similar shape of the traceplot over iterations indicates that the posterior distributions are converging}
    \label{fig_6}
\end{figure}

Fig-\ref{fig_3} shows the ELBO trajectory over iterations for 5000 random initializations. Over iterations it can be observed that the ELBO is converging. This exhibits property of the ELBO that it lower-bounds the log evidence from Equation \ref{eq3}, $\log$ $p(D)$ $\geq$ ELBO for any $q(w|\theta)$ \cite{blei2017variational}.

Fig-\ref{fig_4} shows the comparison of the Receiver Operator Characteristics (ROC) between our BNN classifier and a No-Skill classifier, using the test of volunteer-1. Our No-Skill classifier is an untrained one and it predicts a random class in all cases. The No-Skill classifier at each threshold is represented by a diagonal line from the bottom left to the top right of the plot to the top right. The Area Under Curve (AUC) score represents the skill of a model. The No-Skill classifier always has an AUC score of 0.5, compared to this our BNN classifier has an AUC score of 0.855.  

Fig-\ref{fig_5} shows the trace of the sampling of the posterior weight distributions over iterations. Fig-\ref{fig_6} illustrates the traceplot showing the sample value of the weight for each iteration. The similar shape of the traceplot over iterations indicates that the posterior distributions are converging. The dispersion of the posterior weight distribution indicates that there is uncertainty in the model parameter (weight), therefore this is a representation of epistemic uncertainty. This shows us that model uncertainty does play a role in predicting uncertainty and therefore models which do not take uncertainty into consideration have the risk of generating predictions that are off during real world deployment.

The overall average prediction accuracy for the 30 volunteers is 86.44\%. Going forward, it would be interesting to explore uncertainty from the paradigm of Bayesian Convolutional Neural Networks (CNN). The application of CNNs to the classification of fNIRS signals have been investigated and it performed better relative to classification methods like a Support Vector Machine (SVM) \cite{saadati2019convolutional}. However, it is yet to be examined by taking data and model uncertainty into account using Bayesian statistics.   

\section{Conclusion}
This paper presented a model to carry out classification of an open-access fNIRS dataset, while taking uncertainty into consideration. The dataset consisted of unilateral right and left finger tapping. We used a Bayesian Neural Network model to classify between the two tasks. The average classification accuracy over the data of the 30 volunteers is 86.44\%. We illustrated how the ELBO function of the model converges over iterations. We also illustrated how the posterior weight distribution changes over repeated iterations and the corresponding uncertainty associated with it.  We further generated a ROC curve, which showed our model has an AUC score of 0.855. 
\bibliographystyle{IEEEtran}
\bibliography{bare_conf}

\end{document}